\documentclass{article}

\usepackage[square, numbers]{natbib}
\usepackage[final]{neurips_2023}
\usepackage{epsfig}

\usepackage[utf8]{inputenc} % allow utf-8 input
\usepackage[T1]{fontenc}    % use 8-bit T1 fonts
\usepackage{hyperref}       % hyperlinks
\usepackage{url}            % simple URL typesetting
\usepackage{booktabs}       % professional-quality tables
\usepackage{amsfonts}       % blackboard math symbols
\usepackage{nicefrac}       % compact symbols for 1/2, etc.
\usepackage{microtype}      % microtypography
\usepackage{xcolor}         % colors
\usepackage{graphicx}

\title{Generalization properties of contrastive world models}

\author{%
  Kandan Ramakrishnan\thanks{Department of Neuroscience, Baylor College of Medicine.} \\
  \texttt{Kandan.Ramakrishnan@bcm.edu} \\
  \And
  R. James Cotton \thanks{Shirley Ryan AbilityLab, Northwestern University} \\
  \texttt{rcotton@sralab.org } \\
  \And
  Xaq Pitkow \footnotemark[1] \thanks{Neuroscience Institute, Department of Machine Learning, Carnegie Mellon Univeristy} \thanks{Department of ECE, Rice Univeristy}  \\
  \texttt{xaq@cmu.edu} \\
  \And
  Andreas S. Tolias \footnotemark[1]\\
  \texttt{astolias@bcm.edu} \\
}

\begin{document}

\maketitle

\begin{abstract}
%World models that predict the consequence of actions are presumed to address the key challenge of generalization in current AI systems.
Recent work on object-centric world models aim to factorize representations in terms of objects in a completely unsupervised or self-supervised manner. Such world models are hypothesized to be a key component to address the generalization problem. While self-supervision has shown improved performance  however, OOD generalization has not been systematically and explicitly tested. In this paper, we conduct an extensive study on the generalization properties of contrastive world model. We systematically test the model under a number of different OOD generalization scenarios such as extrapolation to new object attributes, introducing new conjunctions or new attributes. Our experiments show that the contrastive world model fails to generalize under the different OOD tests and the drop in performance depends on the extent to which the samples are OOD. When visualizing the transition updates and convolutional feature maps, we observe that any changes in object attributes (such as previously unseen colors, shapes, or conjunctions of color and shape) breaks down the factorization of object representations. Overall, our work highlights the importance of object-centric representations for generalization and current models are limited in their capacity to learn such representations required for human-level generalization.
\end{abstract}

\section{Introduction}

One of the main challenges in AI is to learn high-level causal variables from low-level variables like pixels in images that enable generalization. Learning world models \cite{huang2020better, karl2016deep, kossen2019structured, veerapaneni2020entity} in a self-supervised manner might be a key component for generalization \cite{lecun2022path}. Such world models also form a core component of human cognition \cite{craik1967nature}. A number of studies in cognitive psychology and neuroscience \cite{spelke1990principles, teglas2011pure, wagemans2015oxford} show that world models aid in generalization and robust learning. For example, infants learn about the physical properties of objects as entities that behave consistently over time and are able to re-apply their knowledge to new scenarios involving previously unseen objects \cite{spelke1995development}. Thus, learning object-centric world models in a self-supervised manner seem to be a key component of cognition that allow humans to predict and generalize to novel situations.

\begin{figure}
    \centerline{\includegraphics[width = 0.7\textwidth]{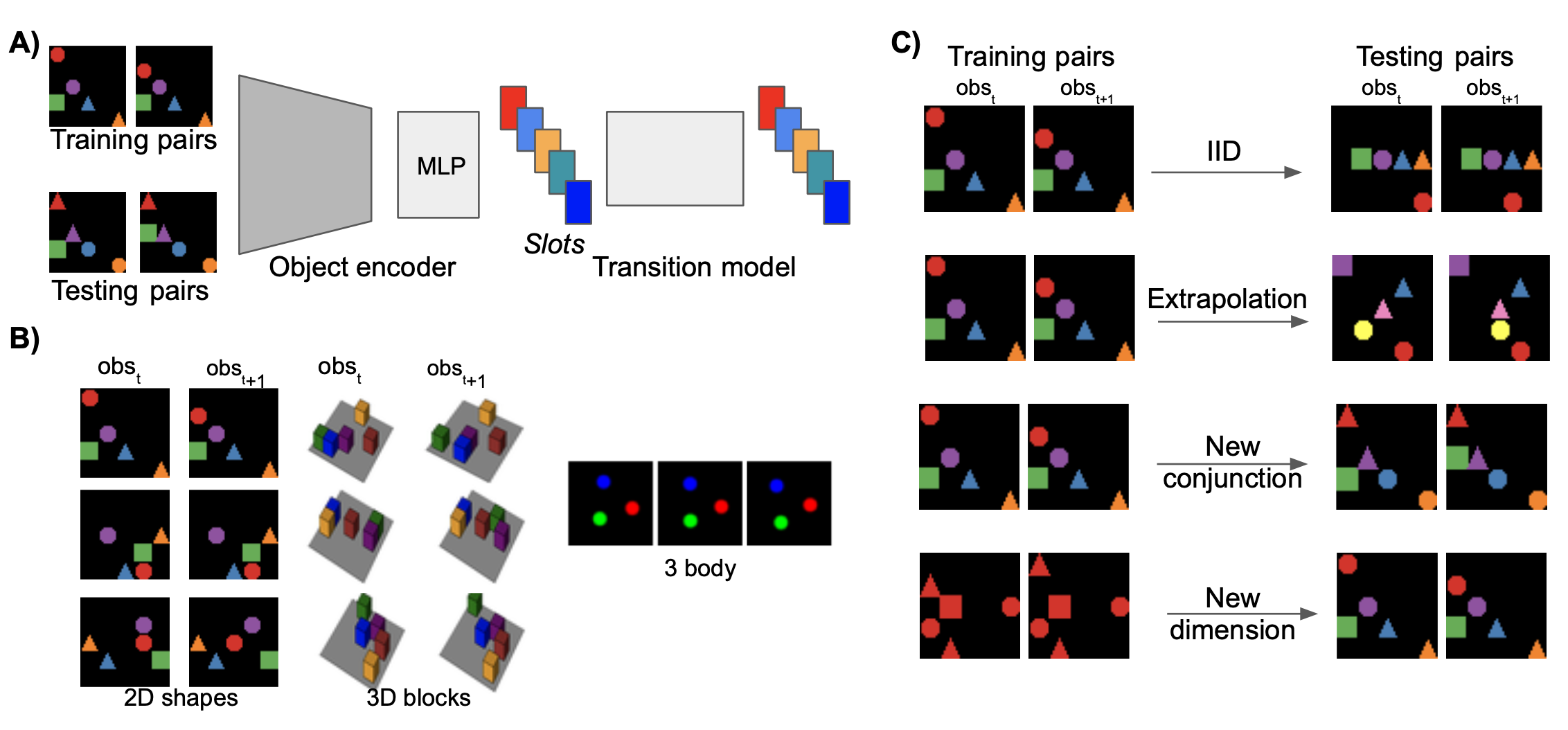}}
    \caption{A) Model architecture of object-centric world model used in our experiments. The world model consists of an object encoder with slot architecture and followed by a Graph Neural Network as the transition model. B) Datasets used to evaluate models for out-of-distribution generalization. The grid based 2D shapes, 3D blocks and 3 body dataset are visualized. C) Illustration of the generalization tests on 2D shapes dataset - i) IID:training and testing data are the same, ii) New conjunction : Testing on novel color-shape combinations not seen during training, iii) Extrapolation : Testing for a new shape or new color different from the training dataset and iv) New dimension - Either variation in shape or color is seen during training while the testing contain variation in both shape and color.}
    \label{fig:short}
\end{figure}

A desirable property of world models is their potential ability to generalize to novel objects as observed in infants \cite{spelke2007core}. While contrastive world models \cite{kipf2019contrastive} have been previously evaluated for their prediction performance, very few studies have tested their effectiveness for generalization to novel data samples. In a recent study \cite{biza2022factored}, world models are shown to have generalization abilities beyond their training environment or number of objects. This study evaluates generalization in terms of single scenario such as novel task and also under different environmental conditions. However in this study it is assumed that the input image is factored into constituent objects by an object detection module. This, however is not a thorough test of generalization for world models. Our aim is to conduct an exhaustive evaluation of the generalization abilities of contrastive world models under a number of different OOD conditions.

In our experiments, we train a contrastive structured world model (CSWM) on a next step prediction task given an input observation and an associated action. The models are trained on different datasets - 2D shapes, 3D blocks and 3-body dataset \cite{jaques2019physics}. The trained models are then tested for different types of generalization. We find that: (1) Object-centric world models are unable to factorize representations under OOD; and (2) the drop in generalization performance depends on the number of time steps and number of objects that are OOD. %and (3) increase in model capacity or using alternate training objective function does not improve generalization performance. 
Overall, our findings challenge the notion that contrastive learning of object-centric world models potentially help with OOD generalization and this requires design of novel learning paradigm to preserve factorization of representations critical for generalization.

\section{Related work}
In \cite{locatello2019challenging} it has been shown that learning disentangled representations requires an inductive bias in the model architecture and the data. Other studies have shown its relation to fairness \cite{locatello2019fairness}, its usefulness for downstream tasks that enable quicker learning from fewer samples \cite{van2019disentangled}. Few studies also study its relation to generalization. \cite{dittadi2020transfer} demonstrates that learning disentangled representations is a good predictor for out-of-distribution task performance. However, in the same study, they also show that VAE-based approaches \cite{locatello2019challenging} do not learn to disentangle complex datasets, which instead requires increased model capacity. In another study, \cite{trauble2021disentangled} investigate the ability of disentangled representations to generalize under OOD conditions. However none of these studies have investigated generalization performance of contrastive world models.

\section{Results}

\subsection{Model evaluation on IID data}

The evaluation of CSWM model on the 2D shapes IID data shows perfect factorization of object-level representations as shown in Figure 2. Given the perfect factorization, CSWM model achieves perfect performance on IID data. 

\begin{figure}
\begin{center}
 \centerline{\includegraphics[width = 0.8\textwidth]{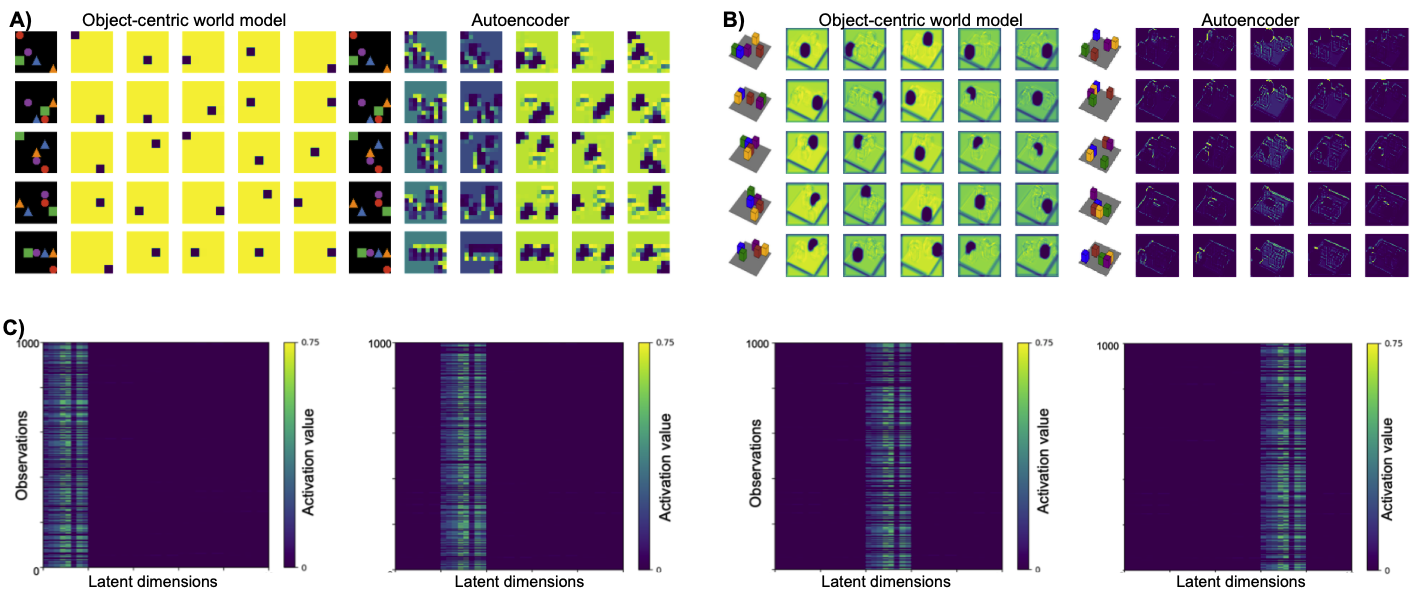}}
\end{center}
   \caption{Understanding the factorization of representations: A) visualization of the  activation maps from the convolutional backbone on 2D shapes dataset of both CSWM and AE. Each map corresponds to an object from the input image which indicates to what extent the encoder is able to factorize the representation space as per objects. B) Visualization of convolutional maps on 3D blocks dataset of both CSWM and AE models. C) Each plot is the state transitions when only one object is moved in the environment.}
\label{fig:exps}
\end{figure}

\subsection{Out-of-distribution evaluation}
To test the hypothesis whether object-centric representations generalize to novel data, we evaluate the models prediction performance under OOD settings by changing the attributes (shape or color) of objects in the dataset. Given that the CSWM outperforms other slot-based models, we evaluate CSWM for OOD prediction performance. For each generalization test, we also vary the number of objects with attributes changed in the test set. 

Figure 3 shows the CSWM performance under different types of generalization on the 2D shapes dataset. We observe that for all types of generalization, the performance deteriorates. The prediction performance depends to what  extent the test samples are OOD: as more the number of objects are changed, the performance worsened. Additionally, we also notice that the performance degrades over longer time-steps. Single-step prediction performance on OOD remains close to IID, especially with only one object changed. However, for higher time steps (5 and 10 time steps) there is a considerable drop in performance. Similarly the CSWM model fails to generalize to 3D blocks and 3 body dataset (Figure 4).

The drop in OOD performance of the model indicates that either the convolution encoder doesn't factorize the representations or that the transition model fails to update the state representation accurately. To identify we run a qualitative analysis on both the encoder representations and the transition updates. As in the IID analysis, we visualize the convolution filters to see if the objects are factored under the OOD conditions. Figure 3B shows the output from the convolution layers of the CSWM object encoder trained on 2D shapes dataset. Unlike in IID, each convolution map does not correspond to a distinct object in the input image. We observe that the convolution map activations mixes mask corresponding to multiple objects. The CSWM model is thus unable to factorize object-level representation under all the different OOD settings. 
Given that the factorization breaks down, the transition model can no longer correctly identify the right object to update its state. This is indeed the case, as shown in Figure 2C. We visualize the transition updates such that for each subplot of observations the same object is acted upon. In OOD conditions, we observe that the transition model updates the state for multiple objects in each observation, even though the true prediction is for the transition model to update the state of just one object in an observation. This is in contrast to IID scenario where we observed that one object state was updated for every observation. The failure in accurate factorization of representations results in inaccurate transition updates as seen in the low prediction performance.

Overall, the slot mechanism of CSWM model is unable to factorize the representations when attributes of the object are changed during test time, resulting in poor prediction performance.

\begin{figure}
\begin{center}
 \centerline{\includegraphics[width = 0.8\textwidth]{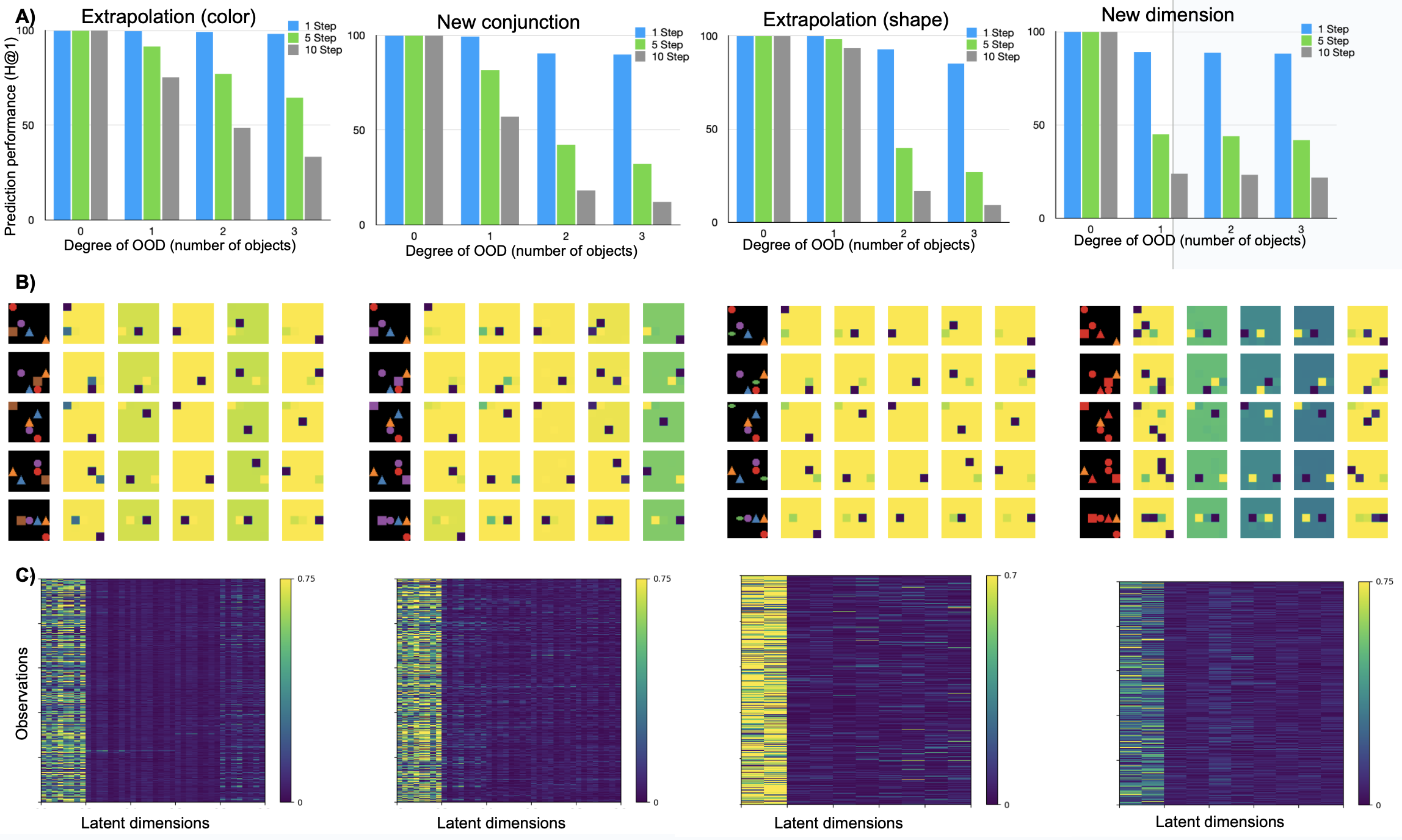}}
\end{center}
   \caption{Evaluation of CSWM model under OOD generalization. A) H@1 prediction performance of the model under different types and extent of OOD. B) Visualization of convolutional maps of the model corresponding to each generalization test. C) Transition updates corresponding to each generalization test.}
\label{fig:exps}
\end{figure}

\begin{figure}
\begin{center}
 \centerline{\includegraphics[width = 0.7\textwidth]{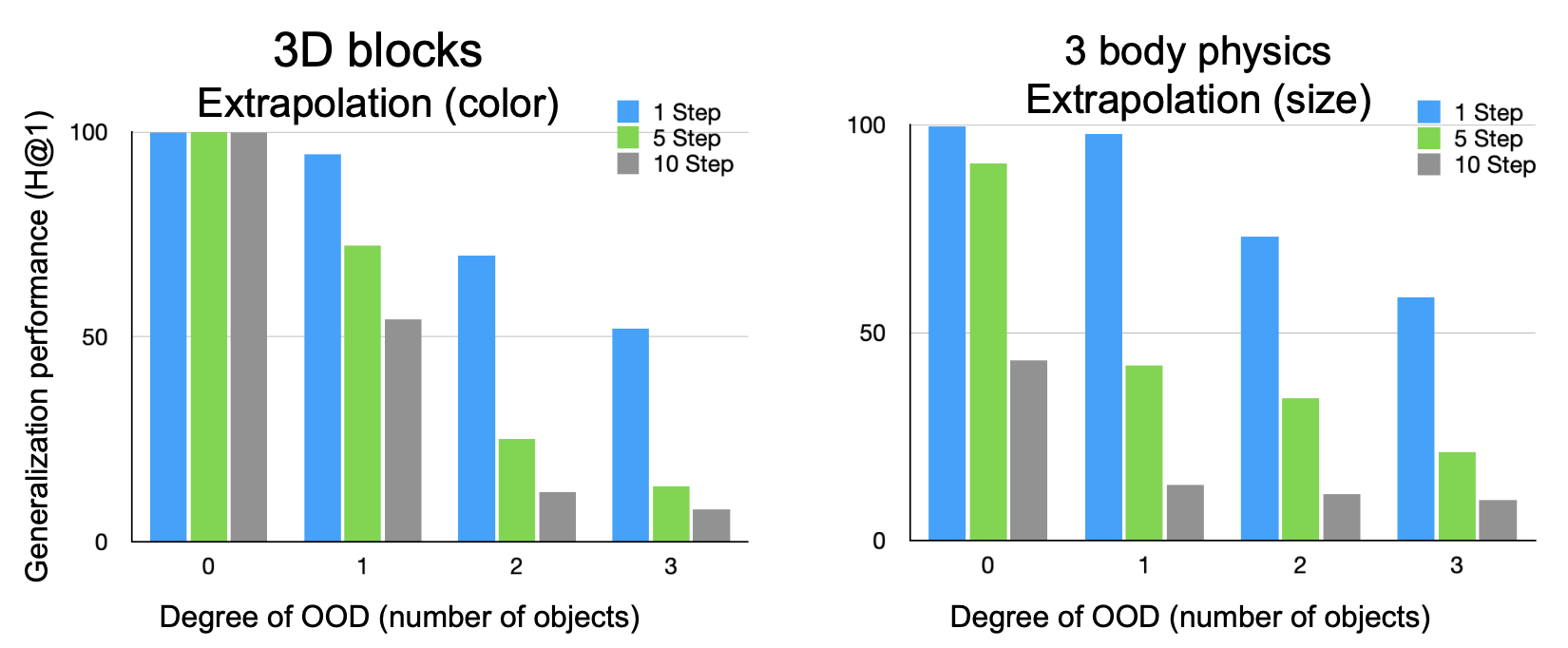}}
\end{center}
   \caption{OOD generalization performance of CSWM on 3D blocks and 3 body dataset.}
\label{fig:exps}
\end{figure}

\section{Discussion}
In this paper, we find that the tested world model is unable to generalize to novel data which can be attributed to the breakdown in factorization of representations. While we aim to conduct an exhaustive evaluation to understand the generalization properties of slot-based models, there are limitations regarding datasets and models. The datasets used in the study are based on synthetic images and have relatively simple object dynamics. As regards to the model, there are other approaches to building object-centric models that we did not test. A number of approaches use generative models \cite{burgess2019monet} to build object-centric representations.

\section*{Acknowledgements}
We would like to thank Zhe Li for helpful discussions. This research has been funded by the NSF NeuroNex program through grant DBI-1707400 and UF1 NS126566 awarded to AST and XP. This work was also supported in part upon work supported by the Air Force Office of Scientific Research (AFOSR) under award number FA9550-21RT0750 to XP.

%%%%%%%%%%%%%%%%%
{\small
\bibliography{egbib}

\begin{thebibliography}{20}
\expandafter\ifx\csname natexlab\endcsname\relax\def\natexlab#1{#1}\fi
\providecommand{\url}[1]{\texttt{#1}}
\providecommand{\href}[2]{#2}
\providecommand{\path}[1]{#1}
\providecommand{\DOIprefix}{doi:}
\providecommand{\ArXivprefix}{arXiv:}
\providecommand{\URLprefix}{URL: }
\providecommand{\Pubmedprefix}{pmid:}
\providecommand{\doi}[1]{\href{http://dx.doi.org/#1}{\path{#1}}}
\providecommand{\Pubmed}[1]{\href{pmid:#1}{\path{#1}}}
\providecommand{\bibinfo}[2]{#2}
\ifx\xfnm\relax \def\xfnm[#1]{\unskip,\space#1}\fi
%Type = Article
\bibitem[{Biza et~al.(2022)Biza, Kipf, Klee, Platt, van~de Meent and
  Wong}]{biza2022factored}
\bibinfo{author}{Biza, O.}, \bibinfo{author}{Kipf, T.}, \bibinfo{author}{Klee,
  D.}, \bibinfo{author}{Platt, R.}, \bibinfo{author}{van~de Meent, J.W.},
  \bibinfo{author}{Wong, L.L.}, \bibinfo{year}{2022}.
\newblock \bibinfo{title}{Factored world models for zero-shot generalization in
  robotic manipulation}.
\newblock \bibinfo{journal}{arXiv preprint arXiv:2202.05333} .
%Type = Article
\bibitem[{Burgess et~al.(2019)Burgess, Matthey, Watters, Kabra, Higgins,
  Botvinick and Lerchner}]{burgess2019monet}
\bibinfo{author}{Burgess, C.P.}, \bibinfo{author}{Matthey, L.},
  \bibinfo{author}{Watters, N.}, \bibinfo{author}{Kabra, R.},
  \bibinfo{author}{Higgins, I.}, \bibinfo{author}{Botvinick, M.},
  \bibinfo{author}{Lerchner, A.}, \bibinfo{year}{2019}.
\newblock \bibinfo{title}{Monet: Unsupervised scene decomposition and
  representation}.
\newblock \bibinfo{journal}{arXiv preprint arXiv:1901.11390} .
%Type = Book
\bibitem[{Craik(1967)}]{craik1967nature}
\bibinfo{author}{Craik, K.J.W.}, \bibinfo{year}{1967}.
\newblock \bibinfo{title}{The nature of explanation}. volume
  \bibinfo{volume}{445}.
\newblock \bibinfo{publisher}{CUP Archive}.
%Type = Article
\bibitem[{Dittadi et~al.(2020)Dittadi, Tr{\"a}uble, Locatello, W{\"u}thrich,
  Agrawal, Winther, Bauer and Sch{\"o}lkopf}]{dittadi2020transfer}
\bibinfo{author}{Dittadi, A.}, \bibinfo{author}{Tr{\"a}uble, F.},
  \bibinfo{author}{Locatello, F.}, \bibinfo{author}{W{\"u}thrich, M.},
  \bibinfo{author}{Agrawal, V.}, \bibinfo{author}{Winther, O.},
  \bibinfo{author}{Bauer, S.}, \bibinfo{author}{Sch{\"o}lkopf, B.},
  \bibinfo{year}{2020}.
\newblock \bibinfo{title}{On the transfer of disentangled representations in
  realistic settings}.
\newblock \bibinfo{journal}{arXiv preprint arXiv:2010.14407} .
%Type = Article
\bibitem[{Huang et~al.(2020)Huang, He, Singh, Zhang, Lim and
  Benson}]{huang2020better}
\bibinfo{author}{Huang, Q.}, \bibinfo{author}{He, H.}, \bibinfo{author}{Singh,
  A.}, \bibinfo{author}{Zhang, Y.}, \bibinfo{author}{Lim, S.N.},
  \bibinfo{author}{Benson, A.R.}, \bibinfo{year}{2020}.
\newblock \bibinfo{title}{Better set representations for relational reasoning}.
\newblock \bibinfo{journal}{Advances in Neural Information Processing Systems}
  \bibinfo{volume}{33}, \bibinfo{pages}{895--905}.
%Type = Article
\bibitem[{Jaques et~al.(2019)Jaques, Burke and Hospedales}]{jaques2019physics}
\bibinfo{author}{Jaques, M.}, \bibinfo{author}{Burke, M.},
  \bibinfo{author}{Hospedales, T.}, \bibinfo{year}{2019}.
\newblock \bibinfo{title}{Physics-as-inverse-graphics: Joint unsupervised
  learning of objects and physics from video}.
\newblock \bibinfo{journal}{arXiv preprint arXiv:1905.11169} .
%Type = Article
\bibitem[{Karl et~al.(2016)Karl, Soelch, Bayer and Van~der
  Smagt}]{karl2016deep}
\bibinfo{author}{Karl, M.}, \bibinfo{author}{Soelch, M.},
  \bibinfo{author}{Bayer, J.}, \bibinfo{author}{Van~der Smagt, P.},
  \bibinfo{year}{2016}.
\newblock \bibinfo{title}{Deep variational bayes filters: Unsupervised learning
  of state space models from raw data}.
\newblock \bibinfo{journal}{arXiv preprint arXiv:1605.06432} .
%Type = Article
\bibitem[{Kipf et~al.(2019)Kipf, Van~der Pol and Welling}]{kipf2019contrastive}
\bibinfo{author}{Kipf, T.}, \bibinfo{author}{Van~der Pol, E.},
  \bibinfo{author}{Welling, M.}, \bibinfo{year}{2019}.
\newblock \bibinfo{title}{Contrastive learning of structured world models}.
\newblock \bibinfo{journal}{arXiv preprint arXiv:1911.12247} .
%Type = Article
\bibitem[{Kossen et~al.(2019)Kossen, Stelzner, Hussing, Voelcker and
  Kersting}]{kossen2019structured}
\bibinfo{author}{Kossen, J.}, \bibinfo{author}{Stelzner, K.},
  \bibinfo{author}{Hussing, M.}, \bibinfo{author}{Voelcker, C.},
  \bibinfo{author}{Kersting, K.}, \bibinfo{year}{2019}.
\newblock \bibinfo{title}{Structured object-aware physics prediction for video
  modeling and planning}.
\newblock \bibinfo{journal}{arXiv preprint arXiv:1910.02425} .
%Type = Article
\bibitem[{LeCun(2022)}]{lecun2022path}
\bibinfo{author}{LeCun, Y.}, \bibinfo{year}{2022}.
\newblock \bibinfo{title}{A path towards autonomous machine intelligence
  version 0.9. 2, 2022-06-27}.
\newblock \bibinfo{journal}{Open Review} \bibinfo{volume}{62}.
%Type = Article
\bibitem[{Locatello et~al.(2019a)Locatello, Abbati, Rainforth, Bauer,
  Sch{\"o}lkopf and Bachem}]{locatello2019fairness}
\bibinfo{author}{Locatello, F.}, \bibinfo{author}{Abbati, G.},
  \bibinfo{author}{Rainforth, T.}, \bibinfo{author}{Bauer, S.},
  \bibinfo{author}{Sch{\"o}lkopf, B.}, \bibinfo{author}{Bachem, O.},
  \bibinfo{year}{2019}a.
\newblock \bibinfo{title}{On the fairness of disentangled representations}.
\newblock \bibinfo{journal}{Advances in neural information processing systems}
  \bibinfo{volume}{32}.
%Type = Inproceedings
\bibitem[{Locatello et~al.(2019b)Locatello, Bauer, Lucic, Raetsch, Gelly,
  Sch{\"o}lkopf and Bachem}]{locatello2019challenging}
\bibinfo{author}{Locatello, F.}, \bibinfo{author}{Bauer, S.},
  \bibinfo{author}{Lucic, M.}, \bibinfo{author}{Raetsch, G.},
  \bibinfo{author}{Gelly, S.}, \bibinfo{author}{Sch{\"o}lkopf, B.},
  \bibinfo{author}{Bachem, O.}, \bibinfo{year}{2019}b.
\newblock \bibinfo{title}{Challenging common assumptions in the unsupervised
  learning of disentangled representations}, in:
  \bibinfo{booktitle}{international conference on machine learning},
  \bibinfo{organization}{PMLR}. pp. \bibinfo{pages}{4114--4124}.
%Type = Article
\bibitem[{Spelke(1990)}]{spelke1990principles}
\bibinfo{author}{Spelke, E.S.}, \bibinfo{year}{1990}.
\newblock \bibinfo{title}{Principles of object perception}.
\newblock \bibinfo{journal}{Cognitive science} \bibinfo{volume}{14},
  \bibinfo{pages}{29--56}.
%Type = Article
\bibitem[{Spelke et~al.(1995)Spelke, Gutheil and Van~de
  Walle}]{spelke1995development}
\bibinfo{author}{Spelke, E.S.}, \bibinfo{author}{Gutheil, G.},
  \bibinfo{author}{Van~de Walle, G.}, \bibinfo{year}{1995}.
\newblock \bibinfo{title}{The development of object perception}.
\newblock \bibinfo{journal}{Visual cognition: An invitation to cognitive
  science} \bibinfo{volume}{2}, \bibinfo{pages}{297--330}.
%Type = Article
\bibitem[{Spelke and Kinzler(2007)}]{spelke2007core}
\bibinfo{author}{Spelke, E.S.}, \bibinfo{author}{Kinzler, K.D.},
  \bibinfo{year}{2007}.
\newblock \bibinfo{title}{Core knowledge}.
\newblock \bibinfo{journal}{Developmental science} \bibinfo{volume}{10},
  \bibinfo{pages}{89--96}.
%Type = Article
\bibitem[{T{\'e}gl{\'a}s et~al.(2011)T{\'e}gl{\'a}s, Vul, Girotto, Gonzalez,
  Tenenbaum and Bonatti}]{teglas2011pure}
\bibinfo{author}{T{\'e}gl{\'a}s, E.}, \bibinfo{author}{Vul, E.},
  \bibinfo{author}{Girotto, V.}, \bibinfo{author}{Gonzalez, M.},
  \bibinfo{author}{Tenenbaum, J.B.}, \bibinfo{author}{Bonatti, L.L.},
  \bibinfo{year}{2011}.
\newblock \bibinfo{title}{Pure reasoning in 12-month-old infants as
  probabilistic inference}.
\newblock \bibinfo{journal}{science} \bibinfo{volume}{332},
  \bibinfo{pages}{1054--1059}.
%Type = Inproceedings
\bibitem[{Tr{\"a}uble et~al.(2021)Tr{\"a}uble, Creager, Kilbertus, Locatello,
  Dittadi, Goyal, Sch{\"o}lkopf and Bauer}]{trauble2021disentangled}
\bibinfo{author}{Tr{\"a}uble, F.}, \bibinfo{author}{Creager, E.},
  \bibinfo{author}{Kilbertus, N.}, \bibinfo{author}{Locatello, F.},
  \bibinfo{author}{Dittadi, A.}, \bibinfo{author}{Goyal, A.},
  \bibinfo{author}{Sch{\"o}lkopf, B.}, \bibinfo{author}{Bauer, S.},
  \bibinfo{year}{2021}.
\newblock \bibinfo{title}{On disentangled representations learned from
  correlated data}, in: \bibinfo{booktitle}{International Conference on Machine
  Learning}, \bibinfo{organization}{PMLR}. pp. \bibinfo{pages}{10401--10412}.
%Type = Article
\bibitem[{Van~Steenkiste et~al.(2019)Van~Steenkiste, Locatello, Schmidhuber and
  Bachem}]{van2019disentangled}
\bibinfo{author}{Van~Steenkiste, S.}, \bibinfo{author}{Locatello, F.},
  \bibinfo{author}{Schmidhuber, J.}, \bibinfo{author}{Bachem, O.},
  \bibinfo{year}{2019}.
\newblock \bibinfo{title}{Are disentangled representations helpful for abstract
  visual reasoning?}
\newblock \bibinfo{journal}{Advances in Neural Information Processing Systems}
  \bibinfo{volume}{32}.
%Type = Inproceedings
\bibitem[{Veerapaneni et~al.(2020)Veerapaneni, Co-Reyes, Chang, Janner, Finn,
  Wu, Tenenbaum and Levine}]{veerapaneni2020entity}
\bibinfo{author}{Veerapaneni, R.}, \bibinfo{author}{Co-Reyes, J.D.},
  \bibinfo{author}{Chang, M.}, \bibinfo{author}{Janner, M.},
  \bibinfo{author}{Finn, C.}, \bibinfo{author}{Wu, J.},
  \bibinfo{author}{Tenenbaum, J.}, \bibinfo{author}{Levine, S.},
  \bibinfo{year}{2020}.
\newblock \bibinfo{title}{Entity abstraction in visual model-based
  reinforcement learning}, in: \bibinfo{booktitle}{Conference on Robot
  Learning}, \bibinfo{organization}{PMLR}. pp. \bibinfo{pages}{1439--1456}.
%Type = Book
\bibitem[{Wagemans(2015)}]{wagemans2015oxford}
\bibinfo{author}{Wagemans, J.}, \bibinfo{year}{2015}.
\newblock \bibinfo{title}{The Oxford handbook of perceptual organization}.
\newblock \bibinfo{publisher}{OUP Oxford}.

\end{thebibliography}
}

\appendix

\section{Experimental details}

In this section, we provide an overview of the setup and experiments we conduct. We first introduce the datasets, model architecture, and evaluation metrics used in the study. Finally, we outline the task and experiments on which the models' predictive performance is evaluated.

\subsection{Data}

Our experiments use three different datasets -- 2D Shape, 3D blocks, and 3 body physics dataset. 2D shapes is a 5×5 grid world consisting of 5 different randomly placed different objects. The overall size of the grid is 50×50. Each object of size 10×10 can only occupy an empty location in the gird and has a unique combination of shape and color. When an action takes places, one object is moved by one position (Fig 1B) and the object cannot be moved if the location is occupied by another object or outside the grid. The 3D blocks dataset is also a block pushing environment, similar to the one used in \cite{kipf2019contrastive}. The rendering component is changed with different perspective and partial occlusions, making it a slightly challenging task. 
The 3-body environment is an interacting system based on classical gravitational dynamics \cite{jaques2019physics} without any actions. The input to the model is two consecutive images of 50×50 that implicitly contains the velocity information. 
%%%%%%%%%% 

To generate an experience buffer for training, we initialize the environment by uniformly sampling objects at random locations. At every time step, we provide state observations as 50 × 50 × 3 tensors with RGB color channels, normalized to [0, 1]. We randomly sample an object-specific action.
Actions are provided as a 4-dimensional one-hot vector (if an action is applied) that encodes the directional movement of a particular object. Else it is represented as a vector of zeros if no action is applied to a particular object. Note that only a single object receives an action per time step. 

\subsection{Model architecture}
The world model as illustrated in Figure 1A comprises an object encoder and a transition model. For the contrastive structured world model (CSWM), we follow the architectural details as given in \cite{kipf2019contrastive}. The object encoder is a convolutional neural network and the transition model is a Graph Neural Network. The transition model accepts a factored latent state and the action vector to predict the next latent state by outputting the transition update as a residual. The network models pairwise interactions between latent state factors (corresponding to objects in the environment) using a fully-connected node network and an edge network. Both are implemented as MLPs with one hidden layer. The edge network outputs an embedding for each directed edge and then aggregated using the node network in order to update the state of each node. Finally, the graph neural network outputs a vector of updated factors.
Finally we also use a standard Autoencoder (AE) in our experiments. 

\subsection{Task evaluation}
We use standard metrics used for evaluating world models directly in latent space. We ask the model to predict the representation of state given an observation and action. The true state representation (obtained from the true observation by taking action in the environment) is compared against the predicted state representation against reference states stored in a buffer. This measure is the Hits at Rank 1 (H@1) and Mean reciprocal rank (MRR). We only report Hits at Rank 1 (H@1) in our experiments.

\subsection{Experiments}
In our experiments we use the contrastive structured world model (CSWM) \cite{kipf2019contrastive} as illustrated in Figure 1A. Our experiments use three different datasets -- 2D Shape, 3D blocks, and 3 body physics dataset. For the OOD evaluation, we test the model under three types of generalization conditions (Figure 1C) i) Extrapolation - object with either a new a color or shape previously unseen during training is introduced ii) New conjunction - object with a unique shape-color combination unseen during training but either the shape or color seen during training iii) New dimension - either shape or color variation is introduced in testing.

\end{document}